\pdfoutput=1
\documentclass[11pt]{article}

\usepackage{emnlp2021}

\usepackage{times}
\usepackage{latexsym}

\usepackage[T1]{fontenc}
\usepackage[utf8]{inputenc}

\usepackage{blindtext}
\usepackage{balance}

\usepackage{afterpage}
\usepackage{multicol}
\usepackage[toc,page]{appendix}

\usepackage{microtype}
\usepackage{comment}
\usepackage{float}
\usepackage{xcolor}
\restylefloat{table}
\usepackage{amsmath}
\usepackage[utf8]{inputenc}
\usepackage{graphicx}

\usepackage{adjustbox}
\usepackage{multirow}
\usepackage{xhfill}
\usepackage{arydshln}

\makeatletter
\def\thickhline{%
  \noalign{\ifnum0=`}\fi\hrule \@height \thickarrayrulewidth \futurelet
   \reserved@a\@xthickhline}
\def\@xthickhline{\ifx\reserved@a\thickhline
               \vskip\doublerulesep
               \vskip-\thickarrayrulewidth
             \fi
      \ifnum0=`{\fi}}
\makeatother

\newlength{\thickarrayrulewidth}
\title{Leveraging Information Bottleneck \\ for Scientific Document Summarization}

\author{Jiaxin Ju$^{1}$, Ming Liu\thanks{$^\ast$Corresponding author} $^{2 3}$, Huan Yee Koh$^1$, Yuan Jin$^1$, Lan Du$^1$ and Shirui Pan$^1$ \\
 $^1$Faculty of Information Technology, Monash University, Australia\\
 $^2$School of Information Technology, Deakin University, Australia\\
 $^3$Zhongtukexin Co. Ltd. , Beijing, China \\
 
 \texttt{\{jjuu0002,hykoh3\}@student.monash.edu}\\
  \texttt{m.liu@deakin.edu.au} \\
  \texttt{\{yuan.jin,lan.du,shirui.pan\}@monash.edu} \\
  }

\begin{document}
\maketitle
\begin{abstract}
This paper presents an unsupervised extractive approach to summarize scientific long documents based on the Information Bottleneck principle. Inspired by previous work which uses the Information Bottleneck principle for sentence compression, we extend it to document level summarization with two separate steps. In the first step, we use signal(s) as queries to retrieve the key content from the source document. Then, a pre-trained language model conducts further sentence search and edit to return the final extracted summaries. Importantly, our work can be flexibly extended to a multi-view framework by different signals. Automatic evaluation on three scientific document datasets verifies the effectiveness of the proposed framework. The further human evaluation suggests that the extracted summaries cover more content aspects than previous systems.
\end{abstract}

\section{Introduction}

Automatic text summarization is a challenging task of condensing the salient information from the source document into a shorter format. Two main categories are typically involved in the text summarization task, one is extractive approach \cite{cheng2016neural,nallapati2016summarunner,xiao2019extractive,cui2020enhancing} which directly extracts salient sentences from the input text as the summary, and the other is abstractive approach \cite{sutskever2014sequence,see2017get,cohan2018discourse,sharma2019entity,zhao2020summpip} which imitates human behaviour to produce new sentences based on the extracted information from the source document. Traditional extractive summarization methods are mostly unsupervised, extracting sentences based on n-grams overlap \cite{nenkova2005impact}, relying on graph-based methods for sentence ranking \cite{mihalcea2004textrank,erkan2004lexrank}, or identifying important sentences with a latent semantic analysis technique \cite{steinberger2004using}. These unsupervised systems have been surpassed by neural-based models \cite{zaheer2020big,huang2021efficient} in respect of performance and popularity, their encoder-decoder structures use either recurrent neural networks \cite{cheng2016neural,nallapati2016abstractive} or Transformer \cite{zhang2019hibert,khandelwal2019sample}. 

\citet{chu2019meansum} developed an unsupervised auto-encoder model which attempts to encode and then reconstructs the documents with some properly designed reconstruction loss. However, as it tries to preserves every detail that helps to reconstruct the original documents, it is not applicable to long-document summarization settings. Recently, \citet{ju2020monash} proposes an unsupervised non-neural approach for long document by building graphs to blend sentences from different text spans and leverage correlations among them. Nevertheless, none of the aforementioned works utilize explicit guidance to aid the model in summarizing a source text. 

To this end, some works \cite{li2018guiding,liu2018generating,zhu2020boosting,saito2020abstractive,dou2021gsum} explore the use of guided signals extracted from the input source document such as keywords, highlighted sentences and others to aid the model architecture in summarizing the input document. These works only utilize a single signal, and \citet{dou2021gsum} empirically showed that if multiple guided signals can be optimally exploited, the model could achieve even greater improvement to its summary outputs in the supervised neural summarization research space. Based on this finding, we propose a multi-view information bottleneck framework that can effectively incorporate multiple guided signals for the scientific document summarization task. 

The original idea of information  bottleneck (IB) principle \cite{tishby2000information} in information theory is to compress a signal under the guidance of another correlated signal. BottleSum \cite{west2019bottlesum} successfully applied IB to the summarization task for short document. Their model generates summary merely by removing words in each sentence while preserving all sentences without considering the importance of the sentences at a document level. It is not suitable for long scientific document summarization as it would preserve all sentences, resulting in significant redundancy. 

\begin{figure}[!t]
    \centering
       \includegraphics[width=0.49\textwidth]{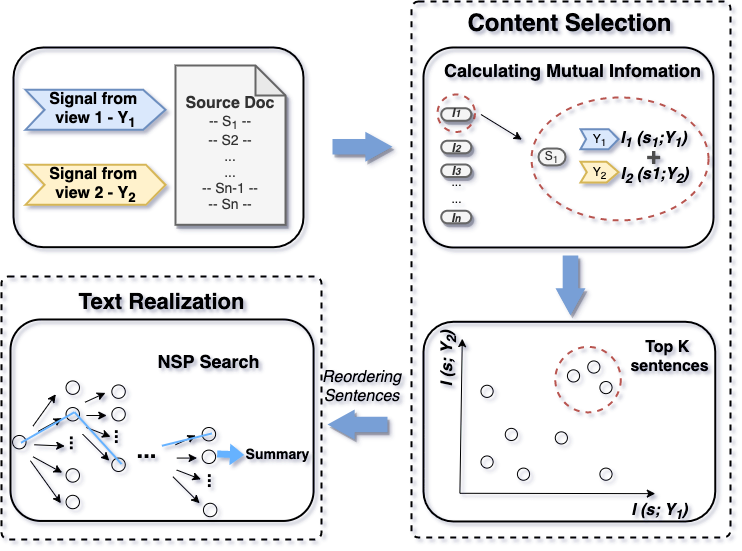}\\
       \caption{Our proposed multi-view information bottleneck framework. $I(s;Y)$ denotes the mutual information between sentence $s$ and correlated signal $Y$, and NSP is short for Next Sentence Prediction task.}
\end{figure}

In contrast, our framework applies IB principle on document level rather than sentence level, where pruning unrelated information will only work on the selected important sentences. In particular, at the content selection stage as shown in Figure 1, the signal that we seek to compress is the source document and the correlated signals are extracted from the source document using state-of-the-art language models. Followed by the text realization step where our proposed architecture conducts sentence search based on fluency to return the final extracted summaries. 

This framework can be flexibly extended to multi-view architecture by incorporating more self-defined correlated signals. Our experiments on arXiv, PubMed \cite{cohan2018discourse} and COVID-19 \cite{wang2020cord19} show that our framework yields competitive performance compared to previous extractive summarizers. Despite the less satisfactory results for multi-view framework in our experiment, we believe it has fruitful potential for further study since the experiments from the work of \citet{dou2021gsum} have empirically shown that summarization through multiple guided signals can achieve significant improvements to the system with a single signal.

\begin{comment}
 Finally, an important work by \cite{dou2021gsum} have explored the use of guided signals in the summarization tasks in the supervised neural summarization research space. Their experiments have empirically shown that summarization through multiple guided signals can achieve significant improvements to current state-of-the-art systems and that no single signal can be dominant in assisting the model results. 
\end{comment}
\section{Information Bottleneck for Unsupervised Summarization}
\subsection{Information Bottleneck principle}

Information bottleneck (IB) principle \cite{tishby2000information} naturally incorporates selection and pruning into compressing information. It compresses the input source $S$ into $\tilde{S}$, which only preserves information related to signal \(Y\), by minimizing the equation:
\begin{equation}
\label{eq1}
I(\tilde{S}; S) - \alpha I(\tilde{S}; Y)
\end{equation} where \(I\) denotes mutual information between two variables and the trade-off coefficient $\alpha$ controls the pruning term and relevance term. The term $I$($\tilde{S}$; $S$) is to prune irrelevant information, while $I$($\tilde{S}$; \(Y\)) enforces the model to retain more information correlated the label \(Y\). For the summarization task, we define $S$ to be the source document, $\tilde{S}$ is the output summary, and $Y$ is the correlated signal. 

\subsection{Multi-view IB for scientific document summarization}

To leverage the benefits of multiple guided signals, we seek to extend IB principles to effectively incorporate multiple guided signals. Recent work \cite{federici2020learning,wu2020phase} leverages the benefit of multi-view IB in other domains, so we extend this framework with multiple views by minimizing the following equation:

\begin{equation}
\label{eq2}
I(\tilde{S}; S) - \alpha \sum_{s \in \tilde{S}}{I(s;Y_1)} - \beta \sum_{s \in \tilde{S}}{I(s;Y_2)}
\end{equation} where $Y_1$ and $Y_2$ refer to two different views on the document content. In this equation, we consider the mutual information between sentences and the guided signal individually. In addition, the term $I$($\tilde{S}$; $S$) still is to remove redundant information, while $\sum_{s \in \tilde{S}}{I(s;Y_1)}$ and $\sum_{s \in \tilde{S}}{I(s;Y_2)}$ are to retain correlated information. The trade-off parameter $\alpha$ and $\beta$ control the relationship between two views $Y_1$ $Y_2$ and the pruning term $I$($\tilde{S}$;$S$) respectively. However, there is no clear way to directly optimize a value for them without a supervised validation set so that we cannot directly compare the importance between two progresses. Instead, we formalise this process as selection step and text editing step. 

Followed by BottleSum\cite{west2019bottlesum}, the equation (\ref{eq2}) can be posed formally as learning a conditional distribution. As we extend their work to document level, the probability of the sentence selected by the system, $P(s)$, should be 1. Then the equation can be formulated as the following:

\begin{equation} 
\label{eq3}
\begin{split}
-logP(\tilde{S}) & - \sum_{s \in \tilde{S}}\Big[\alpha P(Y_1|s)\log P(Y_1|s) \\
& + \beta P(Y_2|s)\log P(Y_2|s)\Big]
\end{split}
\end{equation} Thus, the content selection is to keep relevant information by maximizing $P(Y_1|s)$ and $P(Y_2|s)$ while the text editing step is to prune irrelevant sentences by optimizing $P(\tilde{S})$. In our framework, we define $Y_1$ to be the document categories (e.g. cs, math), and $Y_2$ to be a keyphrase list of the specific article. The equation eventually can be rewritten as\footnote{Detailed derivations of this formula can be found in the Appendix, and our code: https://github.com/Jiaxin-Ju/Unsupervised\_IB\_Summ}:
\begin{equation} 
\label{eq4}
\begin{split}
-logP(\tilde{S}) & - \sum_{s \in \tilde{S}}\Big[\alpha P(Y_1|s)\log P(Y_1|s) \\
& + \beta \sum_{y \in Y_2}P(y|s)\log P(y|s)\Big]
\end{split}
\end{equation} where $y$ is the keywords in the extracted keywords list $Y_2$. Hence, our goal is to maximize $P(Y_1|s)$ and $P(y|s)$ while optimizing $P(\tilde{S})$.

\begin{comment}
We define the guided signal as a list of keywords, and the total mutual information between the generated summary and the guided signal can be factorized upon the summary sentences: $I(\tilde{S};Y) = \sum_{s \in \tilde{S}}I(s;Y)$. Our goal is then to minimize the following equation: 
\end{comment}

\subsection{Proposed algorithm}

To illustrate how our frameworks are learned based on the IB principle, we divide Equation (\ref{eq4}) into two parts and develop an algorithm for each part. The content selection algorithm corresponds to the second term. The algorithm below shows a generalized framework that can be extended to include more than two signals, $Y=\{Y_1, Y_2, ... , Y_n\}$. The implementation details of $Y_i(s)$ will be explained in the section 4.2. The higher the score a sentence gains, the stronger the correlation with the guided signal(s) and the higher the probability it will be included in the output summary. 

\begin{table}
\small
\begin{tabular}{p{7.3cm}}
 \thickhline
 \textbf{Algorithm: Content selection \& Text Realization} \vspace{0.01cm} \\
 \thickhline
 \textbf{Require:} Document $D$, signal set $Y=\{Y_1, Y_2, ... , Y_n\}$, position information $Pos$, and a language model $LM$ \\
 \textbf{Content Selection:} \vspace{0.01cm} \\
 \hspace{0.3cm}1:\hspace{0.1cm}$S_D$ $\leftarrow$ split doc $D$ \hspace{0.5cm} $\ast$ full sentence set \\
% \hspace{0.3cm}2:\hspace{0.1cm}\textbf{for} each $c$ in $C$ \textbf{do}\\
 %\hspace{0.3cm}3:\hspace{0.4cm}$y$ $\leftarrow$ $c(D)$ \hspace{0.5cm} $\ast$ Extract label based on the signal\\
 \hspace{0.3cm}2:\hspace{0.1cm}\textbf{for} each s in $S_D$ \textbf{do}\\
 \hspace{0.3cm}3:\hspace{0.4cm}\textbf{if} len(s) in length constraint \textbf{then}\\
 \hspace{0.3cm}4:\hspace{0.7cm}\textbf{for} each $Y_i$ in $Y$ \textbf{do} \\
 \hspace{0.3cm}5:\hspace{0.99cm}$P(Y_i|s)$ $\leftarrow$ $Y_i(s)$ \\
 \hspace{0.3cm}6:\hspace{0.99cm}$Score(Y_i)$ = $P(Y_i|s) \times log(P(Y_i|s))$ \\
 \hspace{0.3cm}7:\hspace{0.7cm}$Score(s)$ = $\sum_{i=1}^{n}{Score(Y_i)}$ \\
 \hspace{0.3cm}8:\hspace{0.1cm}\textbf{sort} $S_D$ based on $Score(s)$ \textbf{then}\\
 \hspace{0.3cm}9:\hspace{0.4cm}$S_t$ $\leftarrow$ top N sentences in $S_D$\\
 \textbf{Text Realization:} \vspace{0.01cm}\\
 \hspace{0.3cm}1:\hspace{0.2cm}\textbf{sort} $S_t$ based on $Pos$ \textbf{then} \\
 \hspace{0.3cm}2:\hspace{0.2cm}\textbf{for} $m$ in ($0$ ...
 $len(S_t)-1$) \textbf{do} \\
 \hspace{0.3cm}3:\hspace{0.6cm}\textbf{for} $n$ in $m+1$ ... $len(S_t)$ \textbf{do} \\
 \hspace{0.3cm}4:\hspace{1cm} $P_{m,n}$ $\leftarrow$ $LM(S_t[m],S_t[n])$ \\
 \hspace{0.3cm}5:\hspace{0.2cm}matrix $M$ $\leftarrow$ $P_{m,n}$ \\
 \hspace{0.3cm}6:\hspace{0.2cm}\textbf{for} $s$ in $\{first \ k \ sents\}$ of sorted $S_C$\\
 \hspace{0.3cm}\hspace{1.8cm}set $s$ as the start sentence \textbf{do}\\
 \hspace{0.3cm}7:\hspace{0.8cm}best sent path $\leftarrow$ \textbf{Search}($s$, $M$)\\
 \hspace{0.3cm}8:\hspace{0.2cm}best summary path $\leftarrow$ k best sent paths\\
 \hspace{0.3cm}9:\hspace{0.2cm}$S_{sum}$ $\leftarrow$ best summary path\\
 \hspace{0.15cm}10:\hspace{0.2cm}\textbf{return} $S_{sum}$\\
 \thickhline
\end{tabular}
\end{table}

\begin{comment}
To calculate the global score, we use a pre-trained model, Longformer \cite{beltagy2020longformer}, to obtain the $P(Y_G|s)$ for each sentence. Then we interpret the local view as a list of keyphrases that is extracted by RAKE \cite{rose2010automatic}, the local score is the summation of the cosine similarity between the sentence and each keyphrase. Additionally, sentences and keyphrases are mapped into high dimensional space by averaging the output from SciBERT \cite{beltagy2019scibert}, here we assume the sentence with higher similarity to the keyphrases are more likely to associate with the local view. The sentence with a higher score will be selected for the next step.
\end{comment}

For text realization algorithm, the candidate sentence set selected from content selection step is firstly reordered in terms of the sentence original position in the source document. Then we use SciBERT \cite{beltagy2019scibert} to apply the next sentence prediction (NSP) task, each sentence is evaluated against the sentence appeared before it to determine the likelihood that these two sentences are consecutive. Similar idea based on BERT NSP task has been proposed by \cite{bommasani2020intrinsic} to measure a summary's semantic coherence and fluency. Taking fluency of the summary into account, this searching algorithm aims to find the most likely sentence combination as the candidate summary, and the best sentence combination will be selected from these k candidate combinations. Here we implemented greedy search and beam search respectively for model performance comparison. The greedy search algorithm is started from the first sentence, then we find the sentence combination with the highest next sentence probability in each window. For beam search, since the best sentence combination may not start from the first sentence, we perform it for the first k sentences of the candidate sentence set.
%The searching algorithm is performed to find the most likely sentence combination as the candidate summary, and the best sentence combination will be selected from these k candidate combinations.

\begin{table*}[!ht]
\resizebox{\textwidth}{!}{%
\small
\begin{tabular}{p{4.8cm}ccccccccc}
\thickhline 
\thickhline 
  \multirow{2}{*}{}&\multicolumn{3}{c}{\textbf{arXiv}}&\multicolumn{3}{c}{\textbf{PubMed}}&\multicolumn{3}{c}{\textbf{COVID-19}}\\
    &\textbf{R-1}&\textbf{R-2}&\textbf{R-L}&\textbf{R-1}&\textbf{R-2}&\textbf{R-L}&\textbf{R-1}&\textbf{R-2}&\textbf{R-L}\\
%\hline
Oracle &42.37&19.23&39.26&48.10&26.71&45.04&44.61&23.02&40.16\\
Lead-3 &25.68&5.92&20.53&26.71&8.78&23.61&20.88&6.29&18.66\\ [0.5ex]
%Lead-10 &32.38&8.95&25.14&35.43&12.94&29.02&32.00&10.64&24.82\\
\hline
\textbf{Unsupervised Models (rerun)} &&&&&&&&& \\
LSA\cite{steinberger2004using}&32.55&7.54&25.72&34.60&10.07&28.64&30.18&7.12&23.28\\
SumBasic\cite{vanderwende2007beyond} &30.48&6.70&26.34&\textbf{37.76}&11.68&30.19&30.46&7.61&24.63\\
TextRank \cite{barrios2016variations} &31.34&8.68&23.96&33.12&11.84&28.12&27.76&8.60&22.51\\
%LexRank\cite{erkan2004lexrank} &28.47&\textbf{10.85}&\underline{\textit{27.38}}&32.85&\textbf{14.18}&\textbf{32.13}&31.41&\textbf{11.29}&\textbf{26.63}\\
%PACSUM\textsuperscript{$\sharp$} \cite{zheng2019sentence} &\textbf{38.57}&\textbf{10.93}&\textbf{34.33}&\textbf{39.79}&\textbf{14.00}&\textbf{39.31}\\
SciSummPip \cite{ju2020monash} &30.97&\underline{\textit{10.13}}&25.61&36.30&11.96&28.98&29.87&8.27&23.97\\ [0.5ex]
\hline
\textbf{Our Work} &&&&&&&&& \\
keywords only &\underline{\textit{33.40}}&\textbf{10.33}&\textbf{27.66}&\underline{\textit{36.82}}&12.18&29.99&31.61&8.75&24.10 \\
Keywords + beamSearch &\textbf{33.64}& 9.96&\underline{\textit{26.67}}&36.55&\textbf{13.36}&\underline{\textit{30.83}}&\textbf{33.04}&\textbf{9.96}&\underline{\textit{25.22}}\\
keywords + greedySearch &32.13&9.57&26.00&36.19&\underline{\textit{13.19}}&\underline{\textit{30.79}}&\underline{\textit{33.00}}&\underline{\textit{9.87}}&\textbf{25.26}\\
MultiViewIB + beamSearch&32.66&9.05&25.54&36.20&12.98&30.36&32.11&9.32&24.31 \\[0.5ex]
\thickhline
\thickhline
\end{tabular}%
}
\caption{Comparisons with unsupervised extractive models on three scientific datasets. The best F1 results are in \textbf{boldface}, and the second highest scores are in \underline{\textit{italic}}. The implemented TextRank \cite{barrios2016variations} improves the original performance \cite{mihalcea2004textrank} by modifying the similarity function to Okapi BM25.}
\end{table*}
\section{Experiment}
\subsection{Datasets}

\begin{table}
\centering
\small
\begin{tabular}{c|c|cc|cc}
\thickhline
  \multirow{3}{*}{\textbf{Datasets}}&\multirow{3}{*}{\textbf{test docs}}&\multicolumn{2}{c|}{\textbf{Median}}&\multicolumn{2}{c}{\textbf{Median}}\\
    &&\multicolumn{2}{c|}{\textbf{doc length}}&\multicolumn{2}{c}{\textbf{abstract length}}\\
    &&words&sents&words&sents\\
\hline
arXiv &6,440&4319&203&142&6\\
\hline
PubMed &6,658&2293&82&190&7\\
\hline
Covid19 &5,178 &3906&140&231&8\\
\thickhline
\end{tabular}
\caption{Elementary data statistics for the test sets of three datasets. We select approximately number of papers as the COVID-19 test set.}
\end{table}

Additional to the widely-used arXiv and PubMed datasets \cite{cohan2018discourse}, we also make use of the COVID-19 scientific paper dataset \cite{wang2020cord19}. The dataset statistics can be seen in Table 2.

\subsection{Experiment Setup}
\paragraph{Content selection} We define a list of keyphrase extracted by RAKE \cite{rose2010automatic} as the correlated signal for single view, while the multi-view framework incorporates the document category as another view. Top 10 keyphrases are extracted, and sentences and keyphrases are then mapped into high dimensional space by averaging the output from SciBERT \cite{beltagy2019scibert}. We assume the sentence with higher similarity to the keyphrases are more likely to associate with the defined signal, and the score is the summation of the cosine similarity between the sentence and each keyphrase.  For multi-view framework, we use Longformer \cite{beltagy2020longformer} that is pre-trained with 100 classes on the kaggle arXiv dataset\footnote{https://www.kaggle.com/Cornell-University/arxiv} to obtain the $P(Y|s)$ for each sentence. In the pre-training process, we utilize the large model of Longformer and we set the learning rate as 1e-5, batch size as 4, epoch as 4, hidden dropout as 0.05 and the hidden size as 1024. 50 sentences with higher score will be selected for the next step.

\begin{comment}
We pre-train Longformer \cite{beltagy2020longformer} with 100 classes on the kaggle arXiv dataset\footnote{https://www.kaggle.com/Cornell-University/arxiv} that contains 1.7 million articles. After inspecting the dataset, we find that some articles have more than one label and the label distribution is imbalanced. Thus, we remove the multi-labelled documents and randomly sample an approximate equal number of documents for each label. 
\end{comment}

\paragraph{Text Realization} For NSP task, we continue to use SciBERT \cite{beltagy2019scibert} to obtain the likelihood of two adjacent sentences. We implement greedy search and beam search respectively for model performance comparison. The greedy search algorithm is started from the first sentence (k=1), then we find the sentence combination with the highest next sentence probability in each window. We set the window size to 3 then slide the window by one sentence. For beam search, since the best sentence combination may not start from the first sentence, we perform it for the first 5 (k=5) sentences of the candidate sentence set and we set the beam size to 5. The number of sentences in the generated summary is 10.

\begin{comment}
These probabilities are stored in a matrix that has the shape $50 \times 50$, which means we select top 50 sentences with higher score for implementing the searching algorithm. In the beam search, we set the beam size to 5 and we repeat this searching process for the first 5 sentences. While the greedy search is started from the first sentence, and  The number of selected sentence for generated summary is 10.
The searching algorithm is performed to find the most likely sentence combination as the candidate summary, and the best sentence combination will be selected from these k candidate combinations. Here we implement greedy search and beam search respectively for model performance comparison. The greedy search algorithm is started from the first sentence, then we find the sentence combination with the highest next sentence probability in each window. For beam search, since the best sentence combination may not start from the first sentence, we perform it for the first k sentences of the candidate sentence set.
\end{comment}

\subsection{Experiment Results}
\begin{comment}
\paragraph{Oracle and Lead-k} The lower bound of ROUGE score\cite{lin2003automatic} is approximated by taking the first 3 sentences, while the upper bound is determined by using a greedy search method as described in \cite{nallapati2016summarunner}.
\end{comment}

\begin{figure*} [!t]
    \centering
       \includegraphics[width=\textwidth]{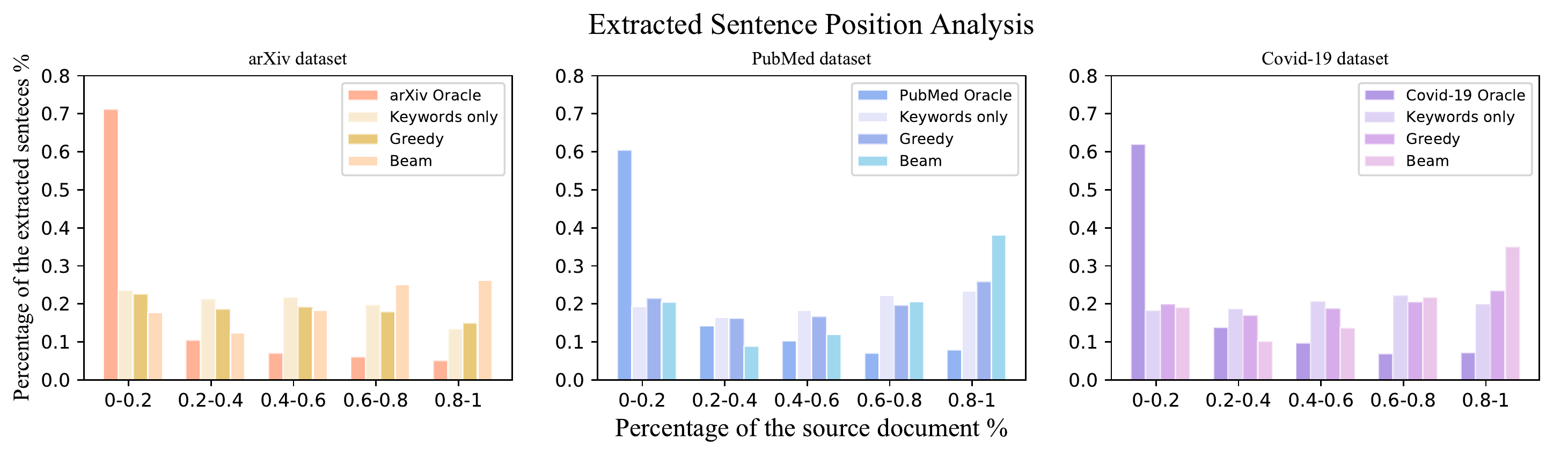}\\
       \caption{The sentence position distribution of the extracted summaries and the oracle summaries.}
\end{figure*}

\paragraph{Results on scientific datasets} We compare our framework with unsupervised summarization models as shown in Table 1. We rerun these models and the number of sentences in the generated summary from all models is 10. Our models achieve the highest R-1 on arXiv and the highest R-2 on PubMed. On the COVID-19, the keywords+beam search setting achieves the highest score. SciSummPip \cite{ju2020monash} is a hybrid method that compresses and rewrites extracted sentences by building a word-relational graph, so it is likely to have more bigrams that match the reference summary. SumBasic \cite{vanderwende2007beyond} tends to extract the sentence that contains more high frequency word so that it achieves a higher R-1 on PubMed. The comparison among our frameworks shows that single view settings performs better than multi-view setting, and the beam search algorithm is better than the greedy search algorithm. While we achieve better scores than baseline results, the performance differences are not significant. Thus, to investigate the effectiveness of our proposed framework, we further conduct the position analysis and human evaluation on single view settings.

\begin{table}
\centering
\resizebox{0.48\textwidth}{!}{%
  \small
  \begin{tabular}{p{3.4cm}cccc}
    \thickhline
    \thickhline
    Model &\textbf{Flu} &\textbf{Fai} &\textbf{Cov} &\textbf{Con}\\
    \thickhline
    Oracle &4.2&4.6&4.0&2.4 \\
    Keywords &2.8&4.0&4.0&3.0 \\
    Keywords + beamSearch &3.4&4.0&4.2&3.4 \\
    Keywords + greedySearch& 3.4&4.2&3.8&3.2\\ [0.5ex]
  \thickhline
  \thickhline
\end{tabular}%
}
\caption{Human Evaluation. Flu, Fai, Cov and Con refer to Fluency, Faithfulness, Coverage and Conciseness.}
\end{table}

\paragraph{Sentence position analysis} Our position analysis is shown in Figure 2., the oracle summaries are mostly extracted from the beginning of the source document, while summaries extracted by our models are from all the sections within the source document. Achieving a higher ROUGE score can prove that the model captures unigram/bigram appeared in the reference summary, but the more important thing is the extracted summary can concisely cover most/all of the key information in each section of the original article for the reader and our model seems to achieve this significantly better than the oracle summary. To prove this hypothesis, we conduct a thorough human analysis. 

\paragraph{Human analysis} We randomly sample 50 documents from the COVID-19 dataset and conduct human evaluation against four criteria: fluency, faithfulness, coverage and conciseness. For each article we compare summaries generated from 4 frameworks with the true summary, and the human annotators are asked to blind rate these summaries on a 1-5 point scale (1 is the worst and 5 is the best). The average performance of each model is shown in Table 3. Even though keywords+beamSearch setting does not significantly perform better than others in terms of ROUGE score, it receive higher human ratings. In addition, oracle summary perform better on fluency and faithfulness but it contains more unnecessary sentences. Figure 3 shows an example\footnote{Original paper can be found at https://arxiv.org/pdf/2007.13933.pdf} of the abstract and the system summaries.

\begin{figure}[!t]
    \centering
       \includegraphics[width=0.48\textwidth]{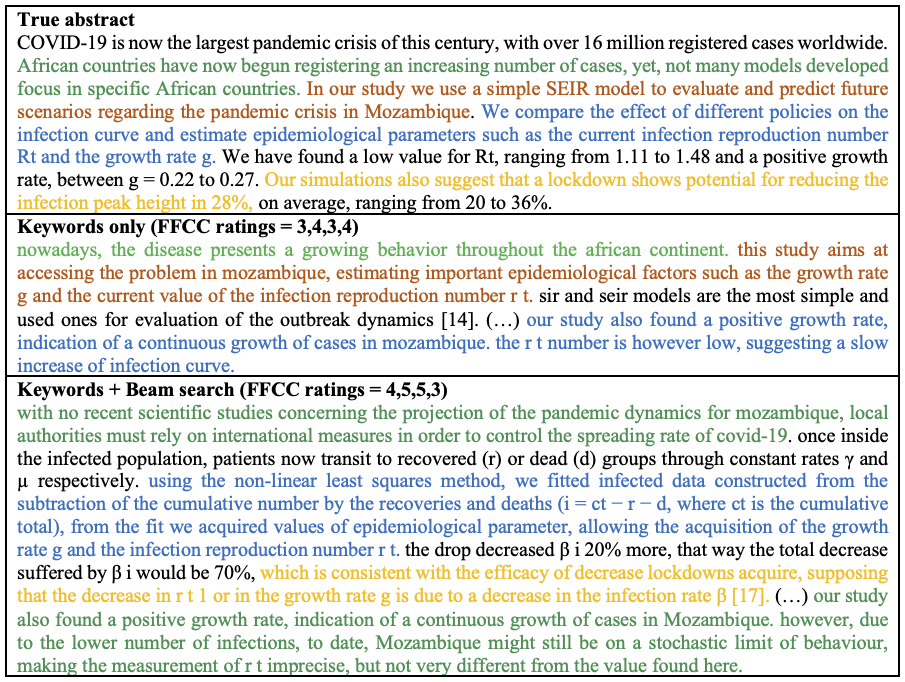}\\
       \caption{An Example taken from COVID-19 dataset. Text in the same color indicates the contents they described are the same.}
\end{figure}

\section{Conclusion and Future work}
In this paper, we proposed an unsupervised framework based on IB principle for long document summarization. Our framework employs a two-steps system where content selection is guided by defined signal(s) and is followed by a text realization step where a pre-trained language model conducts sentence search to return final summaries. Experiments on three scientific show the effectiveness of our framework. Further human analysis suggests that the extracted summaries exhibit more coverage. Despite the less satisfactory results for multi-view framework in our experiment, we believe it has fruitful potential for further study.

% Entries for the entire Anthology, followed by custom entries
\bibliography{anthology,custom}
\bibliographystyle{acl_natbib}

\clearpage
\newpage
\appendix
\label{sec:appendix}

\section{Derivation of formula}

$$I(\tilde{S}; S) - \alpha I(\tilde{S};Y_1) - \beta I(\tilde{S}; Y_2)$$

$$=I(\tilde{S}; S) - \alpha \sum_{s \in \tilde{S}}I(s;Y_1) - \beta \sum_{s \in \tilde{S}}I(s; Y_2)$$

Followed the previous work \cite{west2019bottlesum}, we can rewrite the equation into  an alternate form as shown below:

 $$p(S,\tilde{S})pmi(\tilde{S};S) - \alpha \sum_{s \in \tilde{S}} P(Y_1,s)pmi(s;Y_1) - \beta \sum_{s \in \tilde{S}}p(Y_2,s)pmi(s;Y_2)$$ where $pmi(x,y) = \frac{p(x,y)}{p(x)p(y)}$ denotes pointwise mutual information.

$$=P(S,\tilde{S})\log\frac{P(S,\tilde{S})}{P(S)P(\tilde{S})} - \alpha \sum_{s \in s} P(Y_1,s)\log\frac{P(Y_1,s)}{P(Y_1)P(s)} - \beta \sum_{s \in \tilde{S}}P(Y_2,s)\log\frac{P(Y_2,s)}{P(Y_2)P(s)}$$

$$=P(\tilde{S}|S)P(S)\log\frac{P(\tilde{S}|S)}{P(\tilde{S})} - \alpha \sum_{s \in \tilde{S}} P(Y_1|s)P(s)\log\frac{P(Y_1|s)}{P(Y_1)} - \beta \sum_{s \in \tilde{S}}P(Y_2|s)P(s)\log\frac{P(Y_2|s)}{P(Y_2)}$$

$P(\tilde{S}|S)$=1 for chosen summary, $P(\tilde{S})$, $P(s)$, $P(Y_1)$ and $P(Y_2)$ are constant.

$$=P(\tilde{S}|S)P(S)\log\frac{P(\tilde{S}|S)}{P(\tilde{S})} - \sum_{s \in \tilde{S}}P(s) \left [\alpha P(Y_1|s)\log\frac{P(Y_1|s)}{P(Y_1)} + \beta P(Y_2|s)\log\frac{P(Y_2|s)}{P(Y_2)} \right]$$

$$=C_1log\frac{1}{P(\tilde{S})} - \sum_{s \in \tilde{S}}C_2 \left [\alpha P(Y_1|s)\log P(Y_1|s) + \beta P(Y_2|s)\log P(Y_2|s) - C_3 \right]$$

$-log(p_G)$ is constant, so $P(Y_1|s) \times \log P(Y_1)$ is constant. For each sentence, $P(Y_1|s)$ will be scaled up or down in the same proportion, because $P(Y_1)$ and $P(Y_2)$ are constant and $Log$ is a monotonically increasing function.

$$=-logP(\tilde{S}) - \sum_{s \in \tilde{S}}\left [\alpha P(Y_1|s)\log P(Y_1|s) + \beta P(Y_2|s)\log P(Y_2|s)\right]$$

\end{document}